\documentclass[9pt,conference]{IEEEtran}
\IEEEoverridecommandlockouts
\usepackage{pifont}
\usepackage{cite}
\usepackage{amsmath,amssymb,amsfonts}
\usepackage{algpseudocode}
\usepackage{graphicx}
\usepackage{textcomp}
\usepackage{xcolor}
\usepackage{algorithm}
\def\BibTeX{{\rm B\kern-.05em{\sc i\kern-.025em b}\kern-.08em
    T\kern-.1667em\lower.7ex\hbox{E}\kern-.125emX}}
\begin{document}

\newcommand{\todo}[1]{{\color{red} #1}}
\newcommand{\rev}[1]{{\color{blue} #1}}

\title{Sparse Attention Remapping with Clustering for Efficient LLM Decoding on PIM\\
}

\author{
\IEEEauthorblockN{1\textsuperscript{st} Zehao Fan}
\IEEEauthorblockA{\textit{Electrical, Computer, and Systems Engineering} \\
\textit{Rensselaer Polytechnic Institute}\\
Troy, USA \\
fanz2@rpi.edu}
\and
\IEEEauthorblockN{2\textsuperscript{nd} Garrett Gagnon}
\IEEEauthorblockA{\textit{Electrical, Computer, and Systems Engineering} \\
\textit{Rensselaer Polytechnic Institute}\\
Troy, USA \\
gagnog@rpi.edu}
\and
\IEEEauthorblockN{3\textsuperscript{rd} Zhenyu Liu}
\IEEEauthorblockA{\textit{Electrical, Computer, and Systems Engineering} \\
\textit{Rensselaer Polytechnic Institute}\\
Troy, USA \\
liuz32@rpi.edu}
\and
\IEEEauthorblockN{4\textsuperscript{th} Liu Liu\IEEEauthorrefmark{1}}
\IEEEauthorblockA{\textit{Electrical, Computer, and Systems Engineering} \\
\textit{Rensselaer Polytechnic Institute}\\
Troy, USA \\
liu.liu@rpi.edu}
}

\IEEEoverridecommandlockouts
\IEEEaftertitletext{\vspace{-1.5em}%
\thanks{\IEEEauthorrefmark{1} Corresponding author: liu.liu@rpi.edu}}

\maketitle

\begin{abstract}
Transformer-based models are the foundation of modern machine learning, but their execution, particularly during autoregressive decoding in large language models (LLMs), places significant pressure on memory systems due to frequent memory accesses and growing key-value (KV) caches. This creates a bottleneck in memory bandwidth, especially as context lengths increase. Processing-in-memory (PIM) architectures are a promising solution, offering high internal bandwidth and compute parallelism near memory. However, current PIM designs are primarily optimized for dense attention and struggle with the dynamic, irregular access patterns introduced by modern KV cache sparsity techniques. Consequently, they suffer from workload imbalance, reducing throughput and resource utilization. In this work, we propose STARC, a novel sparsity-optimized data mapping scheme tailored specifically for efficient LLM decoding on PIM architectures. STARC clusters KV pairs by semantic similarity and maps them to contiguous memory regions aligned with PIM bank structures. During decoding, queries retrieve relevant tokens at cluster granularity by matching against precomputed centroids, enabling selective attention and parallel processing without frequent reclustering or data movement overhead. Experiments on the HBM-PIM system show that, compared to common token-wise sparsity methods, STARC reduces attention-layer latency by 19\%--31\% and energy consumption by 19\%--27\%. Under a KV cache budget of 1024, it achieves up to 54\%--74\% latency reduction and 45\%--67\% energy reduction compared to full KV cache retrieval. Meanwhile, STARC maintains model accuracy comparable to state-of-the-art sparse attention methods, demonstrating its effectiveness in enabling efficient and hardware-friendly long-context LLM inference on PIM architectures.
\end{abstract}


\section{Introduction}\label{sec:intro}
Large language models (LLMs) have demonstrated exceptional capabilities across a wide range of natural language processing tasks and are increasingly deployed in real-world applications such as interactive chat systems \cite{achiam2023gpt,zhuang2024toolqa}, code generation tools \cite{svyatkovskiy2019pythia,roziere2023code}, and decision support \cite{shinn2023reflexion,yao2023tree, li2022pre}. However, during inference decoding, LLMs operate in an autoregressive manner, requiring repeated attention computations over a growing sequence of historical key-value (KV) pairs \cite{pope2023efficiently}. As context lengths continue to expand with larger models, the size of the KV cache grows proportionally, resulting in increasingly frequent and large-scale memory accesses. Even with modern hardware support, current GPU architectures typically exhibit a significant imbalance between computational throughput and memory bandwidth, which in turn leads to memory-bound behavior during access-intensive inference workloads \cite{kwon2023efficient}. To address this limitation, processing-in-memory (PIM) technology~\cite{gao2019computedram,newton,hyun2024pathfinding,oliveira2022accelerating,imani2019floatpim}, which integrates computational units within memory systems, emerges as a promising alternative to traditional architectures by alleviating bandwidth constraints and enabling efficient in-memory computation. Recent research has leveraged PIM and heterogeneous computing architectures (e.g., GPU-PIM, NPU-PIM) to accelerate LLM inference \cite{attacc, neupims}. These approaches typically process the memory-bound attention layers using PIM modules, while efficiently handling compute-bound feed-forward networks (FFNs) and Query-Key-Value (QKV) generation with traditional accelerators (xPUs).

However, the trend toward increasingly longer context continues to impose substantial computation and memory costs, primarily due to the quadratic complexity inherent in attention computations. Recent advancements attempt to alleviate this burden through selective retrieval or compression of the KV cache, where only a subset of past tokens' keys and values are selected to approximate full attention computation. However, despite these opportunities, prior works on applying PIM to LLM inference mainly focus on full KV cache attention computations without adapting to emerging sparse attention paradigms. Since state-of-the-art KV cache sparsity methods can achieve sparsity levels above 90\% with minimal impact on model performance, and considering that sparsity dynamically changes during inference decoding, traditional PIM data layouts face challenges due to rigid row-level access granularity. This mismatch leads to workload imbalance in PIM units, reducing throughput and resource utilization. Some approaches, such as Quest~\cite{quest}, mitigate this by retrieving KV pairs at the granularity of pages, each consisting of a fixed number of consecutive tokens. This not only reduces retrieval overhead proportionally by a factor of \(1/\mathit{page\ size}\), but also aligns well with the memory organization of a PIM accelerator. When the page size matches the physical memory row size or is a multiple thereof, the PIM system can fetch and process entire rows efficiently, maximizing memory bandwidth utilization. Nevertheless, since pages are segmented simply by their textual positions, these methods often lead to coarse-grained retrieval. Each retrieved page may include tokens that are irrelevant to the current query, thus wasting computational resources that could be better allocated to tokens with higher semantic importance, degrading model accuracy.

\begin{figure*}[!ht]
  \centering    
  \includegraphics[width=0.9\textwidth]{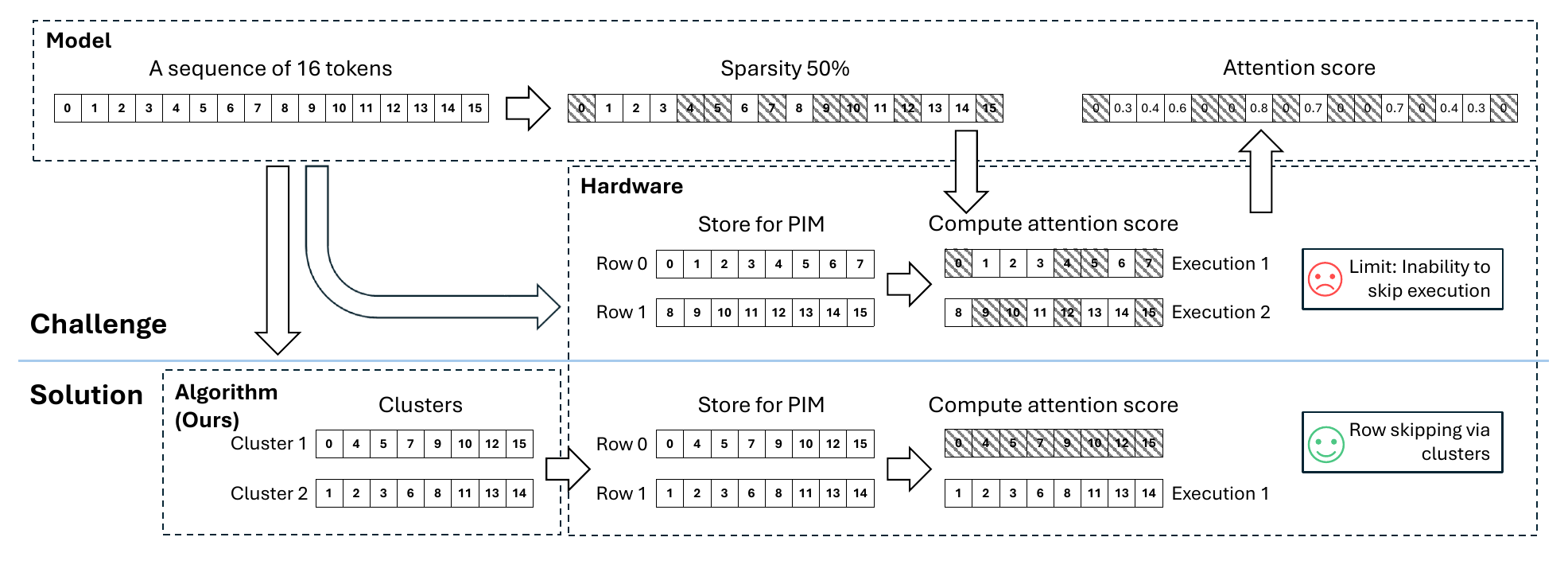}
  \vspace{-15pt}
  \caption{Enhanced execution efficiency through STARC. Due to the coarse granularity of PIM, directly applying sparsity to KV caches often fails to skip computation. STARC addresses this by clustering keys and values such that selected tokens are physically co-located, enabling effective computation skipping and realizing the speedup benefits of sparsity on PIM.}
  \label{fig:motivation1}
\end{figure*}

To bridge this gap, we introduce STARC, a novel sparsity-optimized data mapping scheme tailored specifically for PIM architectures. Fig.~\ref{fig:motivation1} illustrates the key insight behind STARC: by clustering semantically similar tokens and physically co-locating their key-value pairs in memory, the system enables efficient execution under dynamic sparsity while improving overall hardware utilization. Specifically, during the decoding stage, STARC performs online clustering of KV pairs, organizing tokens with high semantic similarity into the same cluster. To facilitate efficient memory access, tokens belonging to the same cluster are proactively placed in contiguous memory locations, aligning with the physical storage layout of PIM banks. When a query arrives, relevant tokens are retrieved at the cluster granularity by comparing it with precomputed cluster centroids, ensuring accurate selective attention while enabling concurrent processing of semantically similar tokens. Notably, once the clusters are generated, they remain unchanged during subsequent decoding steps, thereby avoiding the costly data movement and memory access overhead associated with updating the clustering.

In summary, this paper presents the following contributions:
\begin{itemize}
\item We propose STARC, a sparsity-optimized data mapping scheme for PIM architectures in long-context LLM inference that incorporates clustering of tokens as in KV cache, filling a critical gap in the adoption of sparse attention on PIM architectures.
\item We bridge the mismatch between token-wise sparsity and the row-granularity constraint of PIM memory arrays, enabling efficient row-level computation. Compared with token-wise sparsity methods, STARC reduces attention-layer latency by 19\%--31\% and energy by 19\%--27\%. Under a KV cache budget of 1024, it achieves up to 54\%--74\% latency reduction and 45\%--67\% energy reduction compared to full KV cache retrieval.
\item Compared with hardware-friendly page-wise sparsity methods, we improve retrieval accuracy to a level comparable to token-wise sparsity through remapping more relevant tokens as in KV cache from semantic clustering; thus, STARC can enhance model accuracy and output quality while maintaining compatibility with PIM architectures.
\end{itemize}
\section{Background}

\label{sec:background}

In this section, we first introduce PIM architectures as a solution to the memory bandwidth bottlenecks that arise during decoding. We then describe sparse attention techniques, which alleviate the growing computational and memory access costs associated with long-context sequences, providing basic understandings of our proposed design in Section~\ref{sec:design}.

\subsection{PIM for LLM Attention}

Transformer-based LLMs typically perform inference in two stages: \emph{prefill} and \emph{decoding}. During the prefill stage, the entire input sequence (i.e., the user prompt) is processed in parallel to produce the first output token. In the subsequent decoding stage, tokens are generated one by one in an autoregressive manner, each step being appended to the previous sequence and used as input for the next iteration. This iterative process requires the model to repeatedly read from the KV cache, which stores the key and value vectors projected from all previously generated tokens.

While modern GPUs excel in raw floating-point operations (FLOPs), attention computation during decoding is typically memory-bound, given its low arithmetic intensity and frequent memory accesses to tokens stored in the KV cache. As context lengths grow to hundreds of thousands of tokens, data transfers between GPU compute cores and external memory become a bottleneck. This results in suboptimal resource utilization, as much of computational capacity remains idle while waiting for memory transactions to complete.

\begin{figure}[!ht]
  \includegraphics[width=0.45\textwidth]{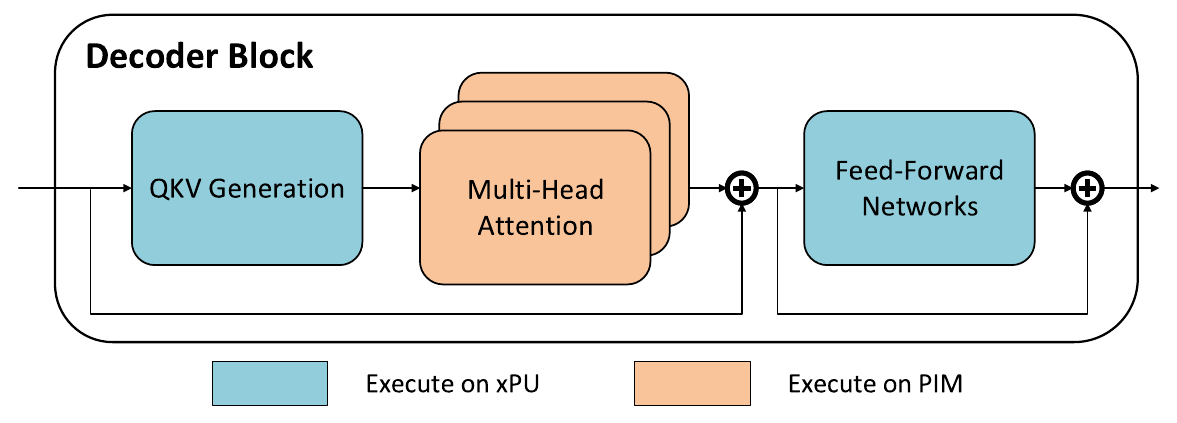}
  \caption{Targeted system overview: the QKV Generation and the Feed-Forward Networks are executed on xPU such as GPU and NPU, while the Multi-Head Attention is executed on PIM.}
  \label{fig:background}
\end{figure}

A more detailed examination of the decoder block architecture provides insight into the source of this memory bottleneck. Each decoder block consists of three fundamental components: (1) \textbf{Query-Key-Value (QKV) generation}, which projects the input hidden states into separate query, key, and value vectors; (2) \textbf{Multi-Head Attention (MHA)}, where attention weights are computed and applied across multiple heads in parallel; and (3) \textbf{Feed-Forward Networks (FFNs)}, which apply independent linear transformations and non-linear activations to each token embedding. Among these components, the MHA module, particularly during the decoding phase, incurs the highest memory bandwidth demand due to its frequent token access to KV cache. Fig.~\ref{fig:background} illustrates the typical execution partitioning adopted by recent PIM-enabled heterogeneous systems, where memory-bound MHA is offloaded to PIM units, while QKV generation and FFNs remain on xPU compute cores.

PIM architectures have emerged as a promising solution to mitigate such memory bandwidth bottlenecks by integrating computation directly within memory systems. Attention layers are especially well-suited for PIM acceleration, primarily for two reasons. First, once the KV matrices for a decoding step are written to the memory arrays, they can be reused repeatedly for subsequent query vectors in the same or following decoding iterations. This reuse pattern allows PIM systems to take full advantage of the high internal bandwidth. Second, MHA operations rely heavily on general matrix-vector multiplication (GEMV) to compute attention scores and outputs. Distributing these computations across parallel memory banks allows the PIM architecture to exploit abundant internal bandwidth while offloading repeated GEMV operations, thereby improving overall throughput under memory-bound conditions.

\subsection{Selective Token Access with Attention Sparsity}

Recent studies on attention distributions in LLMs have revealed that attention scores during inference are often highly sparse. In many cases, only a small subset of tokens significantly contributes to the attention output, while the majority of tokens receive negligible weights. This observation has motivated a range of \emph{sparse attention} techniques that aim to reduce the number of KV pairs accessed during decoding by performing selective retrieval or compression of the KV cache.

\begin{figure}[!ht]
  \centering    
  \includegraphics[width=0.45\textwidth]{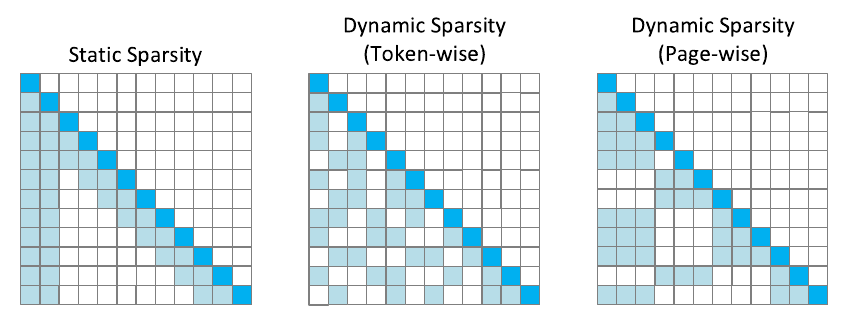}
  \vspace{-5pt}
  \caption{Common attention sparsity patterns.}
  \label{fig:sparsity}
\end{figure}


To be more specific, in the Transformer architecture, each attention head operates on projected query, key, and value vectors. Let \( q \in \mathbb{R}^{1 \times d_h} \) denote the query vector corresponding to the most recent token in a single head, and let \( K, V \in \mathbb{R}^{L \times d_h} \) represent the cached key and value matrices for the \( L \) previous tokens. Here, \( d_h \) denotes the hidden dimension of a single attention head. Sparse attention methods select a subset of size \( B \ll L \), typically based on similarity metrics such as dot-product or cosine similarity, resulting in reduced key and value matrices \( K_S, V_S \in \mathbb{R}^{B \times d_h} \). The attention output is then computed as \( \text{softmax}(qK_S^\top / \sqrt{d_h}) V_S \), where \( q \in \mathbb{R}^{1 \times d_h} \) is the query vector. This selective computation significantly reduces both the memory footprint and the per-step computational cost, while maintaining model quality in many scenarios. Crucially, it also decouples the per-token decoding complexity from the total context length.

In practice, sparse attention mechanisms can be broadly categorized into three representative classes, illustrated in Fig.~\ref{fig:sparsity}:
Firstly, \textbf{static sparsity} restricts each query to attend only to a fixed-size window of past tokens (e.g., the most recent \( B \) tokens), independent of content. This approach is hardware-friendly but fails to capture long-range dependencies.
Secondly, \textbf{dynamic sparsity (token-wise)} selects the top-\( B \) most relevant tokens for each query dynamically based on similarity scores. It provides finer control over which tokens are attended but introduces irregular access patterns. Lastly, \textbf{dynamic sparsity (page-wise)} groups the context into fixed-size segments or “pages” and selects relevant pages rather than individual tokens. Compared with token-wise, this method maintains hardware-friendly access patterns but compromises the effectiveness of per-iteration token access as the retrieval of irrelevant tokens within a page.

\section{Motivation}\label{sec:motivation}
This section motivates our proposed design by analyzing the limitations of existing attention mechanisms on PIM architectures. We first highlight the inefficiencies of dense and token-wise sparsity under PIM’s row-level access granularity, focusing on the mismatch between dynamic token relevance and rigid memory access patterns. We then examine page-wise sparsity, which aligns better with hardware constraints but suffers from low relevance density and reduced attention quality. Finally, we motivate a clustering-based remapping strategy that groups semantically similar tokens into contiguous memory rows, aiming to improve execution efficiency without sacrificing the accuracy of token retrieval.

\subsection{Challenges of Attention on Existing PIM Architectures}

Prior PIM architectures for attention are designed to work with a fully dense KV cache~\cite{attacc, transpim, lol, neupims, papi}, where all past tokens are retained throughout decoding. However, with the long contexts used by modern LLMs, dense attention places heavy demands not only on internal bandwidth but also on the limited computational capacity of PIM architectures. Specifically, the in-situ logic embedded near the memory arrays is typically lightweight and optimized for simple row-wise operations. These resources lack the deep pipelining and wide parallelism of traditional GPU compute units, and are constrained by area and energy budgets within the memory die. Moreover, each attention query must access a large number of stored key-value vectors, which are laid out across many memory rows. In PIM architectures, processing even a single token requires activating entire memory rows, since the logic operates at row granularity. When dense attention forces many such activations per query, the system suffers from frequent row switching and high energy costs due to repeated bitline toggling and row precharging. This behavior severely reduces the efficiency of row-parallel execution across memory banks.

Applying sparse attention methods can potentially alleviate this overhead by accessing only a subset of past tokens. However, applying these methods in the current PIM architectures introduces new challenges. A main issue is that token importance changes dynamically during decoding. A token that is unimportant at one step may become crucial later, so it is impossible to anticipate in advance which tokens will dominate the next attention calculation, limiting the effectiveness of static data placement or scheduling strategies in PIM.

These limitations become especially severe under token-wise sparsity, which requires fine-grained retrieval of individual tokens. As illustrated in Fig.~\ref{fig:motivation1}, such fine-grained access patterns are poorly aligned with the row-level execution granularity of PIM architectures. Each PIM array operates at row granularity: the near-memory logic must activate an entire row, bring all entries onto the bit-lines, and perform computation there. When relevant tokens are scattered across multiple rows, the memory controller is forced to read and process every row individually, leading to substantial over-fetching of irrelevant data and redundant computation.

\begin{figure}[!ht]
  \centering
  \includegraphics[width=0.4\textwidth]{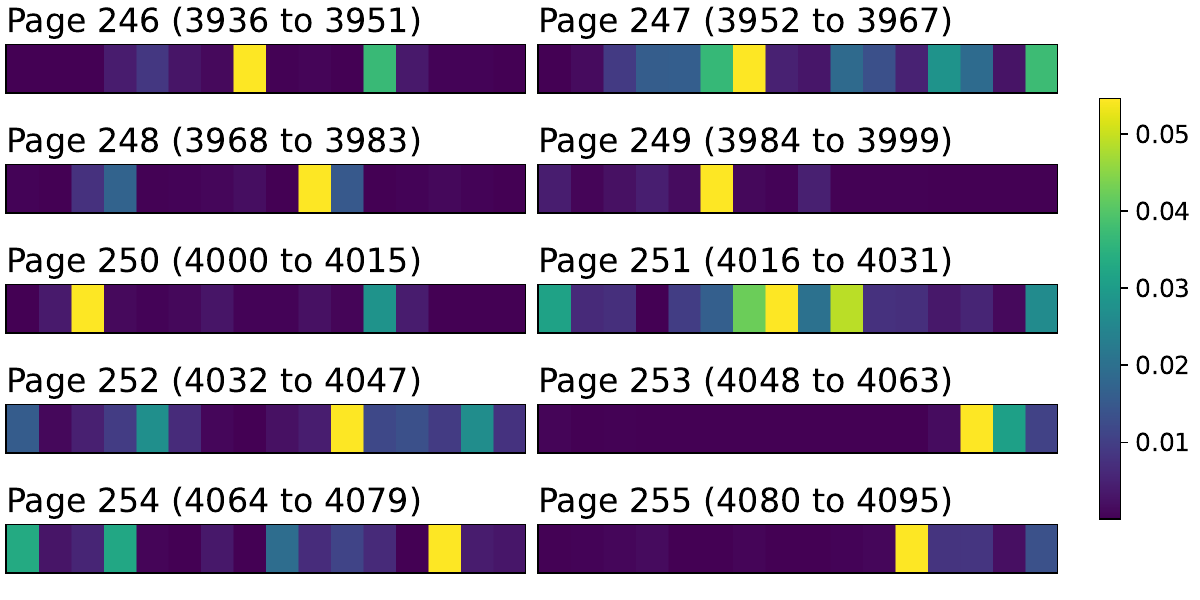}
  \caption{Page-wise retrieval with less important tokens.}
  \label{fig:motivation2}
\end{figure}

\subsection{Hardware Efficiency vs. Attention Quality}

\begin{figure*}[!ht]
  \centering    
  \includegraphics[width=0.9\textwidth]{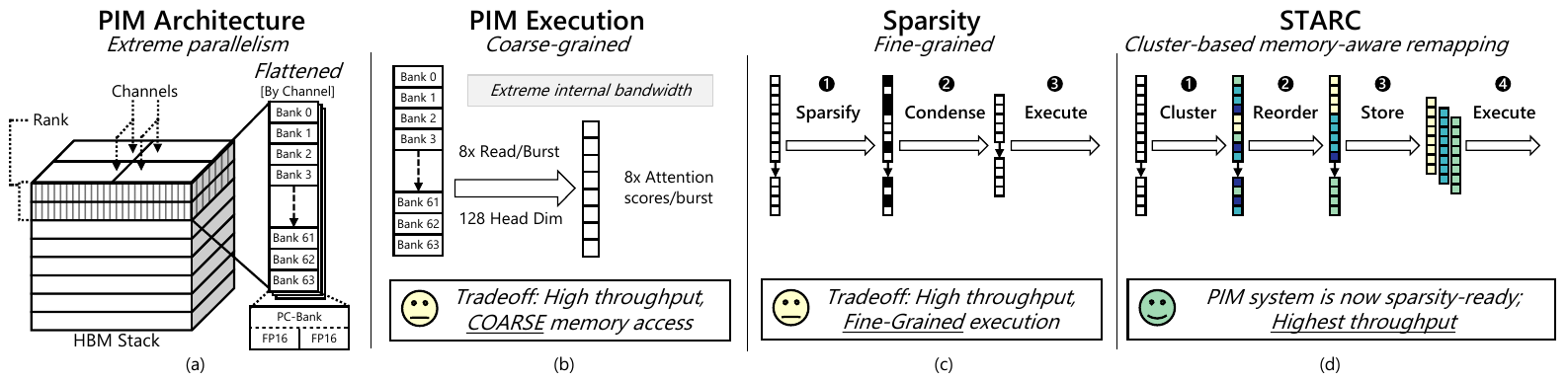}\vspace{-10pt}
  \caption{Supporting sparsity on PIM with STARC. PIM \& sparsity both achieve higher throughput, but enforce vastly different granularities. As a result, the two do not combine readily. STARC applies clustering to improve KV locality in memory, mitigating the granularity required by sparsity while executing with PIM.}
  \label{fig:pim-overview}
\end{figure*}

To reduce the overhead of selecting important tokens during decoding, several recent attention sparsity methods, such as Quest, adopt a page-wise retrieval strategy. In this approach, tokens are grouped into fixed-size pages, and attention is computed over selected pages rather than individual tokens. Quest estimates the criticality of each page by comparing the query vector with the minimal and maximal key vectors of that page, and retrieves only the most relevant ones. This simplifies the sparsity decision process and reduces the complexity of token scoring.

Page-wise token access also aligns well with the memory organization of a PIM accelerator. When the page size is a multiple of the physical memory row size, PIM can fetch and process entire rows efficiently. This allows the accelerator to fully utilize internal memory bandwidth and avoid partial-row access overhead. In HBM-PIM architectures, where computation occurs near DRAM banks, this alignment improves data locality and reduces unnecessary data movement.

However, this hardware compatibility comes at the cost of attention quality and model accuracy. Page boundaries are defined purely based on token position, not on token relevance. As a result, selected pages often include many irrelevant tokens. These tokens are still accessed and processed, wasting bank-level bandwidth and compute resources.

We illustrate this issue using attention heatmaps of LLaMA3.1-8B with a context length of 4K in Fig. \ref{fig:motivation2}. The example uses Quest's page-wise method with a page size of 16 tokens. In the heatmaps, lighter cells represent tokens with higher attention weights. As shown, most pages contain only one or two important tokens. This inefficiency limits the usefulness of page-wise sparsity, despite its compatibility with PIM architectures.

\subsection{Motivation for Remapping and Clustering}



As illustrated in Fig.~\ref{fig:motivation1}, we propose a remapping strategy based on clustering semantically similar tokens, i.e., key and value vectors, and placing each cluster into adjacent, contiguous rows. By aligning row-level memory layout with attention relevance, attention computation in PIM can operate more efficiently: activating a single memory row brings in multiple highly relevant tokens, thereby reducing redundant row activations and unnecessary computation. Compared to conventional sequential mapping, our clustered layout allows each row access to retrieve more meaningful data and enables coarse-grained execution skipping.

This approach aims to achieve the best of two cases. On the one hand, PIM architectures favor regular access patterns that match the row granularity of memory arrays. On the other hand, we want to improve the effectiveness of selective token access by ensuring that each accessed row contains more relevant tokens.
\section{STARC System Architecture}\label{sec:design}

To enable high-throughput execution of attention mechanisms in Transformer-based models, we adopt AttAcc~\cite{attacc} as our PIM architecture---a PIM system specifically designed to accelerate the attention layer. As illustrated in Fig.~\ref{fig:pim-overview}(a), AttAcc places compute units near each bank within an HBM stack. Specifically, each channel comprises 2 ranks, 2 pseudo-channels, 4 bank groups per rank, and 4 banks per bank group, resulting in 64 banks per channel. This configuration supports synchronizing an entire channel’s bandwidth for parallel execution across banks. Thus, when processing, it achieves extreme throughput beyond conventional HBM3.

However, this performance comes at the cost of limited flexibility, as shown in Fig.~\ref{fig:pim-overview}(b). Commands are broadcast across entire channels, forcing all banks within the channel to activate the same row simultaneously. With HBM3’s 8x burst length, each memory read yields 1024 FP16 values, equivalent to processing eight entire key/value vectors when the attention head dimension is 128 (typical for LLaMA-style models). Consequently, PIM execution imposes a coarse granularity: even if only one token is needed, the system must fetch and compute all eight vectors stored within the accessed row.

This granularity presents a critical inefficiency in sparse attention. As shown in Fig.~\ref{fig:pim-overview}(c), sparsity methods typically involve three steps: \ding{202} selecting important tokens, \ding{203} condensing the KV cache, and \ding{204} computing over selected entries. Token-wise sparsity often leads to unstructured selections, where top tokens are distributed across distant memory rows. Since memory access granularity is fixed, the system must still compute all entries in each accessed row, resulting in significant over-fetching and wasted computation.

To address this inefficiency, STARC introduces a clustering-based remapping mechanism that aligns logical sparsity patterns with physical memory layout (see Fig.~\ref{fig:pim-overview}(d)). The key idea is to co-locate tokens that are likely to be sparsified together. Since most sparsity methods select tokens based on query-key similarity, STARC first performs clustering over keys to group semantically similar tokens~\ding{202}. These clusters are then reordered~\ding{203} and written into memory contiguously~\ding{204}, ensuring that each DRAM row contains tokens with a high likelihood of being selected together. During execution~\ding{205}, STARC activates full rows, but with minimal waste, as most or all of the stored entries contribute meaningfully to the computation.

Through this remapping strategy, STARC transforms irregular sparse access into structured, hardware-aligned computation, enabling the PIM system to fully exploit its bandwidth advantages without incurring sparsity-related inefficiencies.

\section{Algorithm Design}

\begin{figure}[!ht]
  \centering    
  \includegraphics[width=0.495\textwidth]{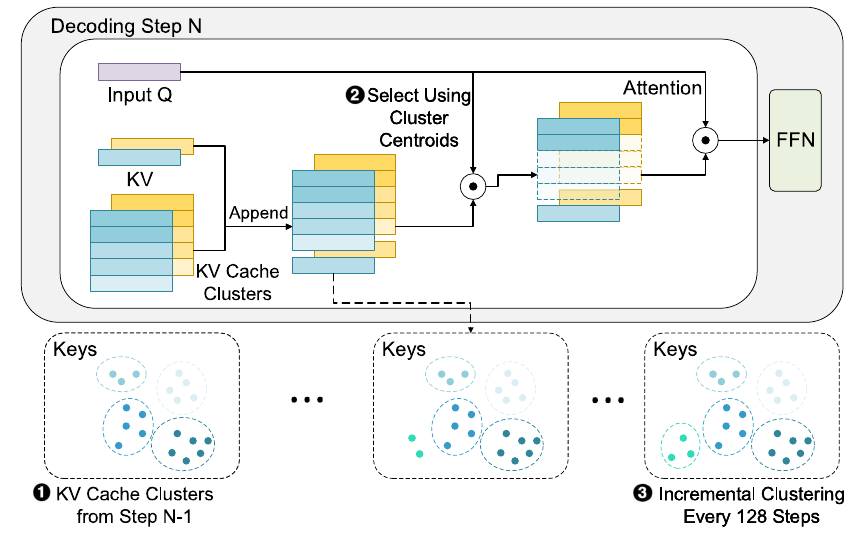} \vspace{-10pt}
  \caption{Flowchart of the clustering algorithm. We perform incremental clustering on the KV pairs using K-means, meaning that only the newly generated segment of KV pairs is clustered during the decoding stage.}
  \label{fig:algorithm}
\end{figure}

\begin{algorithm}[t]
\caption{Clustering-Based KV Retrieval during Decoding}
\label{alg:decoding_cluster}
\begin{algorithmic}[1]
\Require Prefill KV pairs: $\mathcal{K}_\text{pre}, \mathcal{V}_\text{pre}$; 
\\ \hspace{1em} Decoding stream $\{x_t\}$; Clustering interval $I$; KV cache budget $B$
\Statex

\State // \textbf{Initial clustering after prefill}
\State $\mathcal{C} \gets \text{KMeans}(\mathcal{K}_\text{pre})$ 
\Comment{cosine similarity, k-means++ init}
\For{each $(k_i, v_i) \in (\mathcal{K}_\text{pre}, \mathcal{V}_\text{pre})$}
    \State Assign $k_i$ to cluster $\mathcal{C}_j$, and assign $v_i$ to the same $\mathcal{C}_j$
\EndFor

\Statex
\State Initialize: $\mathcal{K}_\text{new} \gets \emptyset$, $\mathcal{V}_\text{new} \gets \emptyset$
\For{each decoding step $t$}
    \State Generate token $x_t$, compute key $k_t$ and value $v_t$
    \State Append $k_t$ to $\mathcal{K}_\text{new}$, $v_t$ to $\mathcal{V}_\text{new}$

    \If{$|\mathcal{K}_\text{new}| = I$}
        \State $\mathcal{C}_\text{new} \gets \text{KMeans}(\mathcal{K}_\text{new})$
        \For{each $(k_i, v_i) \in (\mathcal{K}_\text{new}, \mathcal{V}_\text{new})$}
            \State Assign $k_i$ to cluster $\mathcal{C}_j$, and assign $v_i$ to the same $\mathcal{C}_j$
        \EndFor
        \State $\mathcal{C} \gets \mathcal{C} \cup \mathcal{C}_\text{new}$
        \State Reset $\mathcal{K}_\text{new}, \mathcal{V}_\text{new} \gets \emptyset$
    \EndIf

    \Statex
    \State // \textbf{KV retrieval for current step}
    \State Compute scores $s_j = q_t^\top \mu_j$ for all centroids $\mu_j \in \mathcal{C}$
    \State Sort clusters by $s_j$ in descending order
    \State Select top clusters until total token count reaches $B$
    \State Truncate final cluster if needed to fit budget $B$
    \State Include all non-clustered tokens in $\mathcal{K}_\text{new}, \mathcal{V}_\text{new}$ in attention
\EndFor
\end{algorithmic}
\end{algorithm}

Building upon the STARC framework described above, we introduce an online clustering strategy that incrementally reorganizes tokens in the KV cache throughout decoding. The goal is to reconcile attention accuracy with AttAcc’s row-level access granularity by grouping semantically similar tokens into hardware-aware clusters that serve as the fundamental access units. Although the clusters may not always align perfectly with a single HBM row, the regularized access pattern they impose substantially curtails row overfetch, lowers internal data movement, and improves effective bandwidth utilization across AttAcc’s parallel banks. The overall procedure is summarized in Algorithm~\ref{alg:decoding_cluster}.

As shown in Fig. \ref{fig:algorithm}, at the beginning of decoding phase (decoding step N is 0), we perform our first clustering step immediately (Step \ding{202}). We apply the standard K-means algorithm to all KV pairs generated during the prefill phase, using cosine similarity as the distance metric and K-means++ for centroid initialization to improve clustering stability. To limit runtime overhead, we set the number of iterations to 15. Empirically, we find that 15 iterations are sufficient for convergence in this setting, and increasing the number of iterations yields negligible improvement in clustering quality. Note that clustering is performed on key vectors only, and the associated Value vectors are assigned to the same clusters as corresponding keys.

During each decoding step, newly generated tokens are temporarily retained in full during each decoding step, as they have not yet been assigned to any cluster. These recent tokens are crucial for accurate generation due to their strong influence on attention distribution. Once the KV pairs are organized into clusters, attention selection during decoding proceeds as follows: the query vector of the current token is compared with each cluster centroid via dot product (Step \ding{203}). Clusters are then ranked based on the attention scores and the top-ranked clusters are selected sequentially until the total number of included tokens reaches the predefined KV cache budget. Since the number of tokens in each cluster is not fixed, the final selected cluster may be partially truncated to ensure the total number of recalled tokens does not exceed the budget. This strategy enables selective attention over the most relevant token groups, effectively limiting memory usage and reducing computational costs.

\begin{figure}[!ht]
  \centering    
  \includegraphics[width=0.45\textwidth]{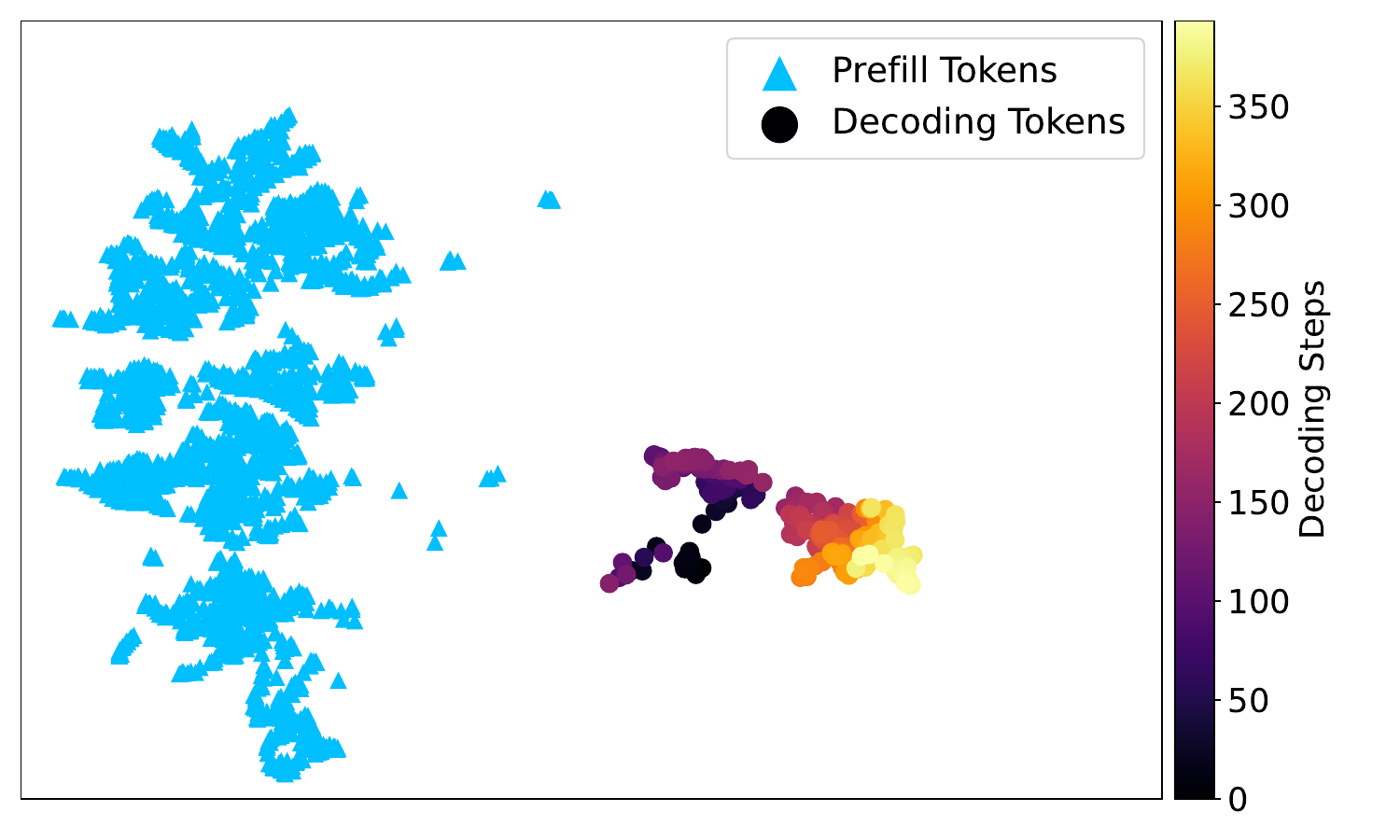}\vspace{-10pt}
  \caption{The distributions of key vectors differ significantly between the prefill and decoding stages.}
  \label{fig:key_distribution}
\end{figure}

In the subsequent decoding phase, we incrementally apply clustering every 128 decoding steps, targeting only the newly generated tokens since the previous clustering (Step \ding{204}). Unlike some dynamic clustering strategies, we do not revisit or update the clustering results of earlier tokens. This design is guided by three key observations. First, the distribution of key vectors generated during decoding gradually diverges from that of the prefill key vectors as decoding progresses, as illustrated in Fig.~\ref{fig:key_distribution}. This motivates us to cluster the prefill and decoding tokens separately in order to maintain semantic consistency within each cluster. Second, key vectors exhibit strong locality, where adjacent tokens often share similar semantics. Prior work~\cite{model} has shown that key vectors of adjacent tokens tend to exhibit high cosine similarity. Leveraging this property, we choose to cluster only the most recent decoding tokens rather than the entire past sequence. This not only improves clustering quality but also reduces latency, making the approach viable for online inference. Third, updating or modifying previously formed clusters would require costly memory access and data movement, especially under PIM’s row-level access granularity. To avoid such inefficiencies, we adopt a growing clustering structure in which new clusters are appended without disrupting existing ones.
\begin{figure*}[!ht]
  \centering    
  \includegraphics[width=0.85\textwidth]{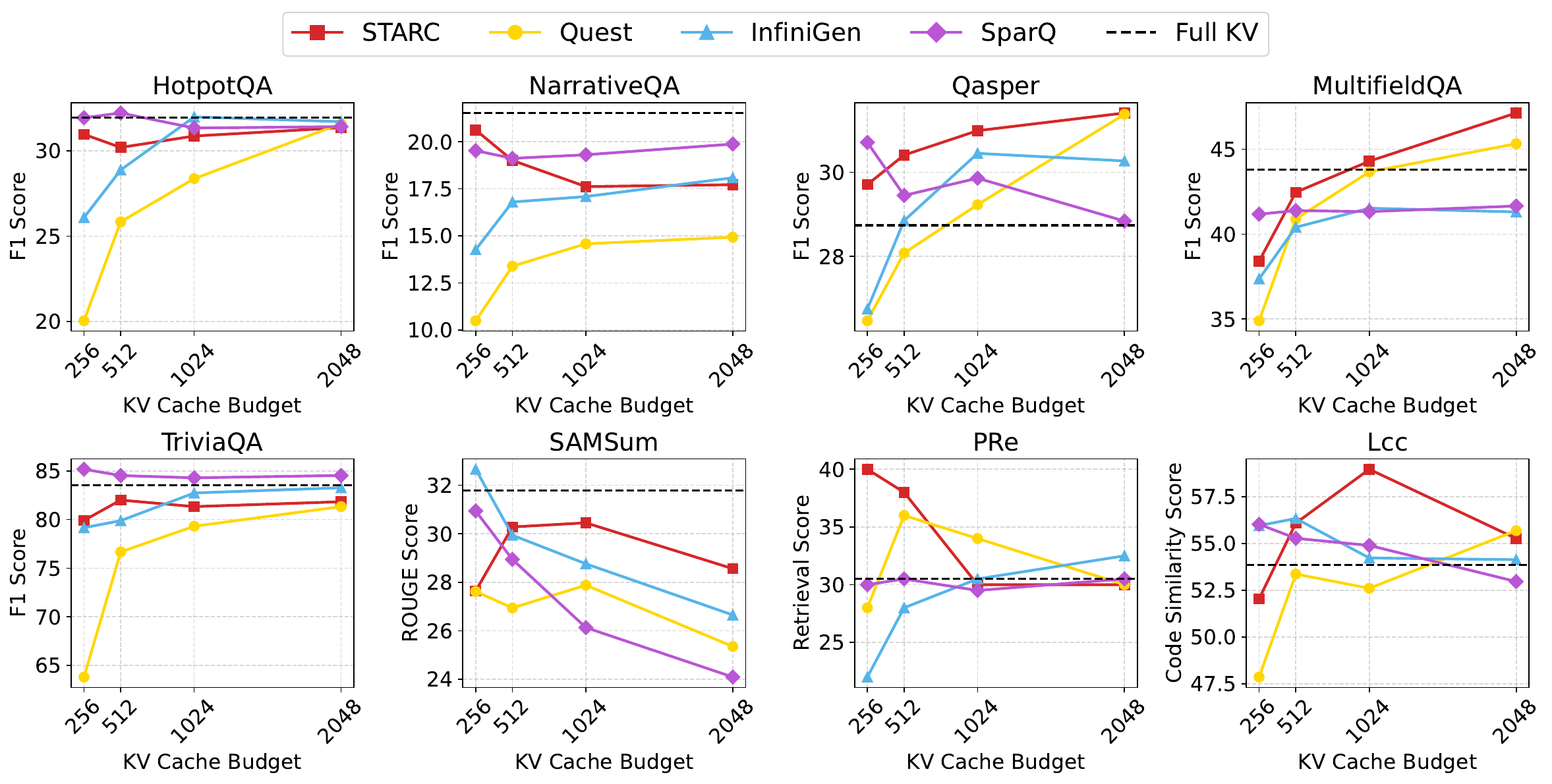} \vspace{-5pt}
  \caption{Results on LongBench of different sparsity methods.}
  \label{fig:longbench}
\end{figure*}

The number of clusters at each step is set to the sequence length divided by 32. This choice reflects a tradeoff between selection granularity and hardware-aware design. From the hardware perspective, setting each cluster to contain approximately 32 tokens is well aligned with the PIM architecture discussed in Section~\ref{sec:design}. This configuration improves row-level utilization and helps distribute memory access more evenly across parallel compute units, mitigating performance bottlenecks caused by workload imbalance. From the algorithm perspective, while increasing the number of clusters can enhance attention quality by enabling finer-grained selection, it also significantly increases the cost of clustering, as the time complexity of K-means grows with the number of centroids. Setting the cluster count proportional to sequence length thus provides a practical balance between attention quality and real-time clustering efficiency.

\section{Evaluation}

\label{sec:eval}

\subsection{Evaluation Methodology}

\paragraph{Accuracy Evaluation} To evaluate the effectiveness of STARC under long-context scenarios, we use LongChat-7B-v1.5-32K \cite{longchat2023} as our base model, a widely used model with 32k context window. For benchmarking, we select eight representative datasets from LongBench~\cite{bai2024longbench}, covering a diverse range of tasks: HotpotQA\cite{yang2018hotpotqa} (multi-document QA), QASPER \cite{dasigi2021dataset}, MultiFieldQA-en, and NarrativeQA \cite{kovcisky2018narrativeqa} (single-document QA), TriviaQA \cite{joshi2017triviaqa} and SAMSum \cite{gliwa2019samsum} (few-shot learning), as well as PRe \cite{raffel2020exploring} (a synthetic task) and Lcc \cite{guo2023longcoder} (code completion). Additionally, we evaluate on PG-19 \cite{rae2019compressive} for language modeling using perplexity as the evaluation metric. We compare our algorithm against three recent KV cache sparsity methods: Quest\cite{quest}, InfiniGen\cite{infinigen}, and SparQ~\cite{sparq}. Each method’s config follows the setting in its original paper (e.g., page size for Quest, partial weights and threshold for InfiniGen, and the largest components \( r \) for SparQ). For a fair comparison, we reproduce all baselines within the same framework. Following the Quest setting, we disable retrieval in the first two layers of the model and instead use the full KV cache, as these layers typically exhibit low sparsity~\cite{quest}. All methods are evaluated under a consistent token budget ranging from 2048 to 256. For our algorithm, we use cosine-based K-means clustering with KMeans++ initialization, and re-cluster every 128 decoding steps. Clustering is performed using Scikit-learn’s K-means implementation. 

\paragraph{Performance on PIM Systems}
To investigate how attention sparsity impacts PIM architectures and evaluate the effectiveness of STARC, we adopt the AttAcc simulator~\cite{attacc}, which extends Ramulator to model heterogeneous GPU–PIM systems, and evaluate on a DGX+AttAccs platform where attention kernels are offloaded to PIM units while FC layers remain on GPU. The DGX consists of 8 NVIDIA H100 cores and 40 HBM3 stacks (5.2\,Gbps per pin), with a total memory capacity of 1.28\,TB. The AttAcc side contains an additional 40 HBM3 stacks, also totaling 1.28\,TB. Each DRAM bank integrates one GEMV unit (1P1B configuration), and all arithmetic and buffer components follow the microarchitectural assumptions in AttAcc~\cite{attacc}.

We compare STARC against full KV cache and two categories of sparsity baselines: token-wise methods, represented by InfiniGen and SparQ, and a page-wise method, represented by Quest. We adopt AttAcc’s optimal configuration by enabling both head-level pipelining and feedforward co-processing, two key architectural optimizations proposed in the original design, to realize its peak achievable performance on attention workloads. To highlight the performance bottleneck introduced by the attention layer during the decoding stage, we configure three sequence length pairs for the prefill and decoding stages: (2k, 2k), (2k, 4k), and (2k, 8k). The target model is LLaMA-7B with FP16 precision. All remaining simulator settings, including arithmetic timing, SRAM/DRAM modeling, datapath lengths, and energy estimation, follow the default AttAcc configuration.

\subsection{Accuracy Evaluation}
\label{sec:accuracy}
\begin{figure}[t]
  \centering    
  \includegraphics[width=0.48\textwidth]{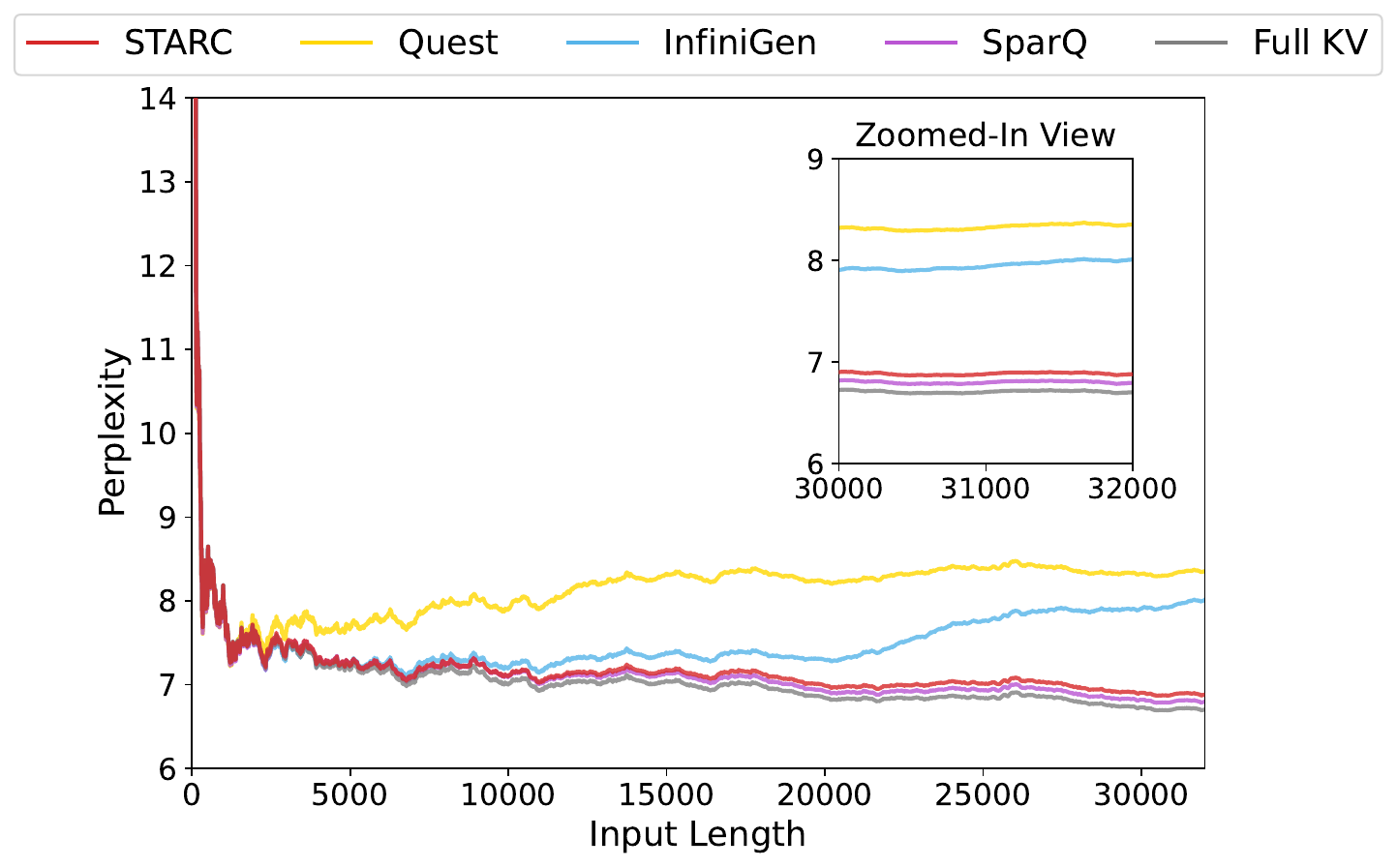} \vspace{-5pt}
  \caption{Language modeling on PG19 dataset.}
  \label{fig:perplexity_comparison}
\end{figure}

\begin{figure}[t]
  \centering
  \includegraphics[width=0.48\textwidth]{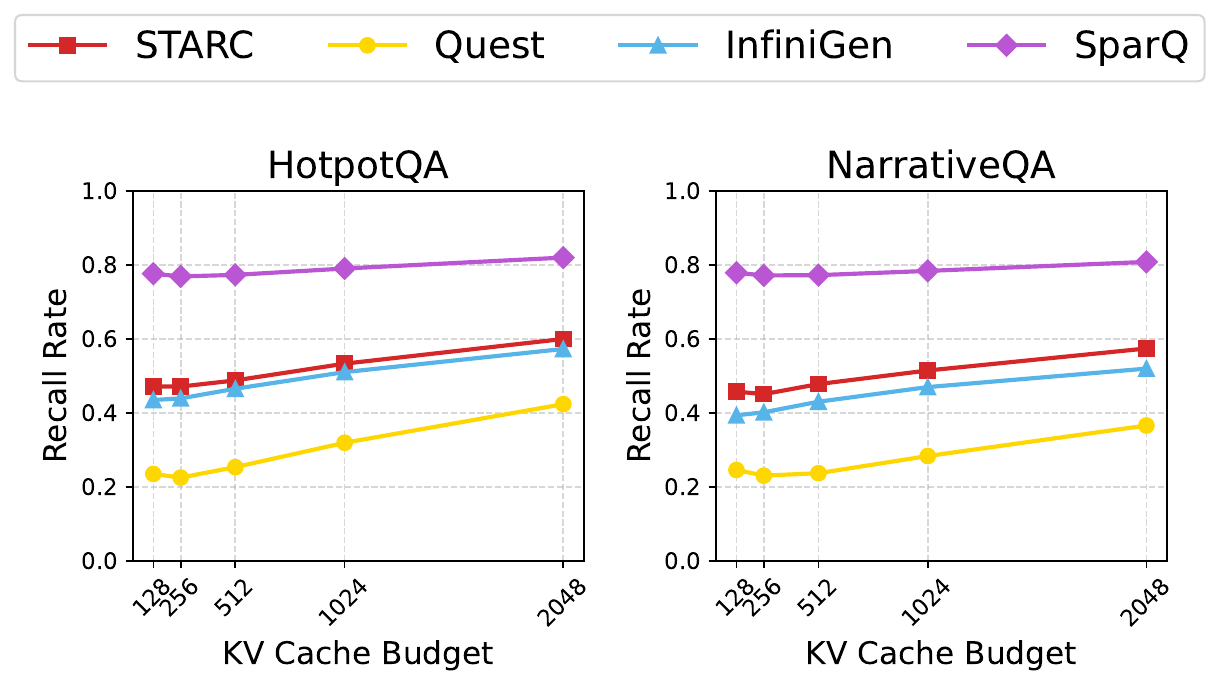} \vspace{-5pt}
  \caption{Recall rate of important tokens.}
  \label{fig:recall}
\end{figure}
\paragraph{Results on LongBench} Fig.~\ref{fig:longbench} presents the results of STARC and baseline methods on the LongBench benchmark. Across all eight datasets and KV cache budgets, STARC consistently outperforms the page-wise sparsity method Quest. It also achieves comparable accuracy to the token-wise sparsity methods InfiniGen and SparQ, and performs better on several datasets, including QASPER, MultiFieldQA, and SAMSum, under most budget settings. These results demonstrate that STARC maintains competitive accuracy compared to state-of-the-art sparsity methods, while offering better hardware alignment with PIM architectures.

\paragraph{Results on Language Modeling} Fig.~\ref{fig:perplexity_comparison} shows the perplexity of generated tokens on the PG-19 test set across varying input lengths, ranging from 1 to 32,000 tokens, under a fixed KV budget of 1024. STARC outperforms both Quest and InfiniGen, particularly at longer input lengths. Although it is slightly outperformed by SparQ, STARC closely follows the Full-KV baseline throughout.

\paragraph{Recall Rate of Important Tokens}  
To intuitively and quantitatively assess how well STARC retrieves important tokens, we extract three samples from the HotpotQA and NarrativeQA datasets respectively and compute the average recall rates for different methods. Given a fixed KV cache budget \(B\), the recall rate is defined as the fraction of the retrieved tokens that fall within the top-\(B\) tokens with the highest attention weights at each decoding step. As shown in Fig.~\ref{fig:recall}, although STARC is outperformed by SparQ, it achieves a higher recall rate than both Quest and InfiniGen. These results indicate that STARC's clustering strategy effectively improves its ability to retrieve semantically important tokens.

\subsection{Performance on PIM Systems}

We evaluate the impact of attention sparsity on the performance of PIM systems, focusing on end-to-end decoding latency and energy efficiency. We categorize sparsity strategies into two types: token-wise sparsity and page-wise sparsity. Token-wise sparsity includes methods such as InfiniGen and SparQ, which generate accurate yet near-random attention masks that vary across decoding steps. Page-wise sparsity is represented by Quest, which selects fixed-size token pages based on position. All methods are evaluated under a consistent KV cache budget of 1024 tokens, which directly determines the number of tokens to be fetched and processed at each decoding step by the PIM simulator.

To assess the hardware efficiency of each method, we first analyze their generated attention masks, which specify the retrieved tokens per decoding step. These masks are compared against the row-level memory granularity of the PIM architecture to compute the actual number of memory accesses required. In AttAcc PIM system shown in Section~\ref{sec:design}, each memory access fetches 8 complete key/value vectors in parallel. This allows us to determine the number of tokens that must be fetched and processed by the simulator at each decoding step from the attention mask.

In the case of Quest, the page size is 16 tokens, which aligns well with the PIM system's row granularity. As a result, there is no over-fetching, and the total number of processed tokens equals the fixed KV budget of 1024. In contrast, token-wise sparsity methods such as InfiniGen and SparQ often retrieve tokens scattered across many rows, leading to additional memory accesses and the processing of irrelevant data. STARC, on the other hand, retrieves tokens at the cluster level. Because semantically similar tokens are stored in the same or adjacent rows during cluster construction, STARC significantly reduces redundant memory activations and over-fetching.

We feed the measured token-level access traces into the AttAcc simulator and evaluate normalized energy and execution time per output token under three sequence configurations: (2048, 2048), (2048, 4096), and (2048, 8192), where the first value indicates the prefill length and the second the decoding length. For reference, we also evaluate the performance of retrieving the full KV cache and normalize all results against this baseline.

\begin{figure}[!ht]
  \centering    
  \includegraphics[width=0.495\textwidth]{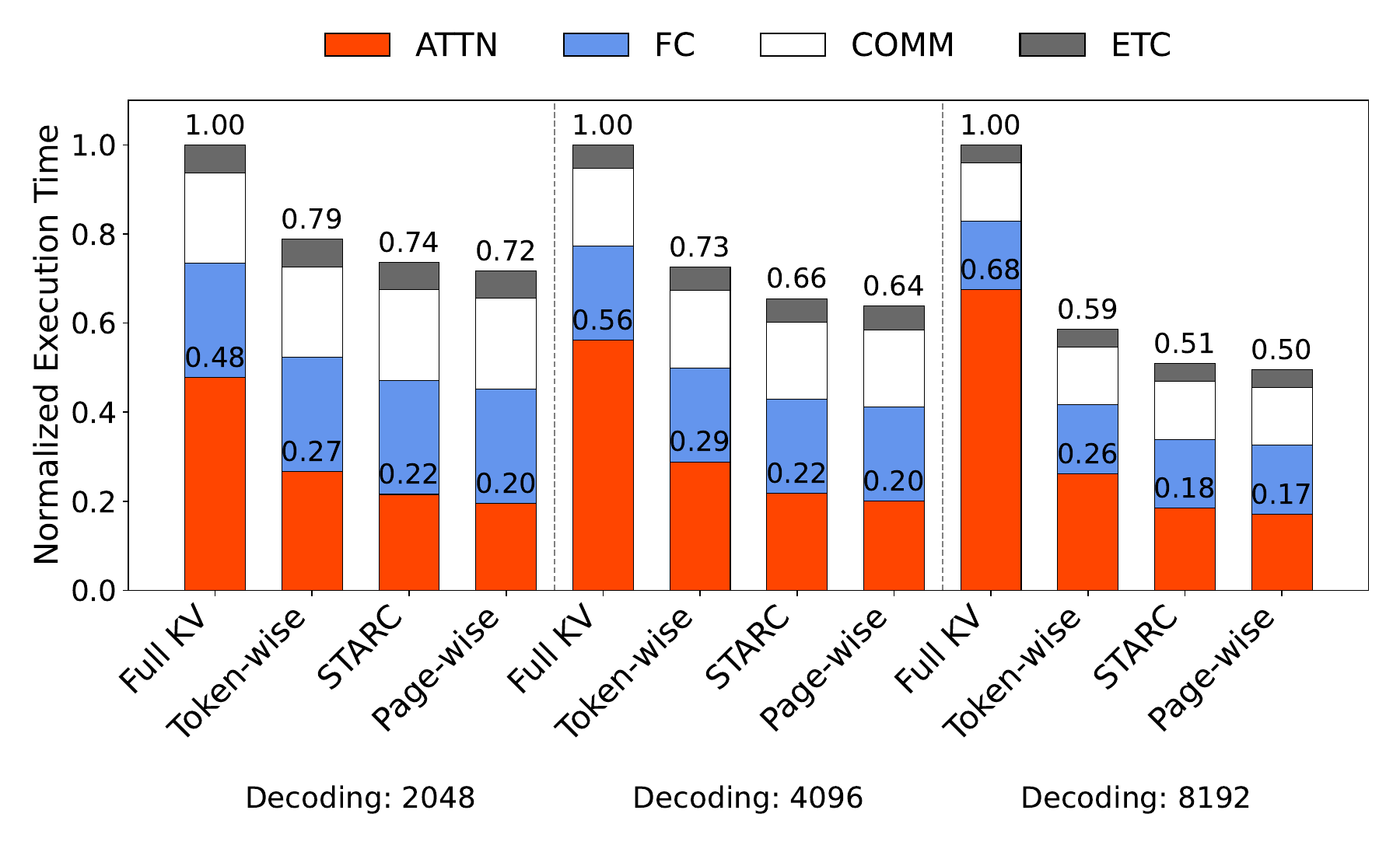} \vspace{-10pt}
  \caption{The normalized execution time per output token of LLAMA 7B for various decoding length.}
  \label{fig:Execution_Time}
\end{figure}

Fig.~\ref{fig:Execution_Time} shows the normalized execution time per token. As decoding length increases, the attention layer becomes the dominant contributor to system latency. Notably, even token-wise sparsity provides up to 40\% speedup over full KV retrieval. Compared to token-wise sparsity methods, STARC reduces attention-layer latency by 19\%--31\%, and achieves up to 54\%--74\% latency reduction relative to full KV retrieval. These improvements translate to 6\%--14\% reduction in total token processing latency, approaching the ideal case of page-wise sparsity.

\begin{figure}[!ht]
  \centering    
  \includegraphics[width=0.495\textwidth]{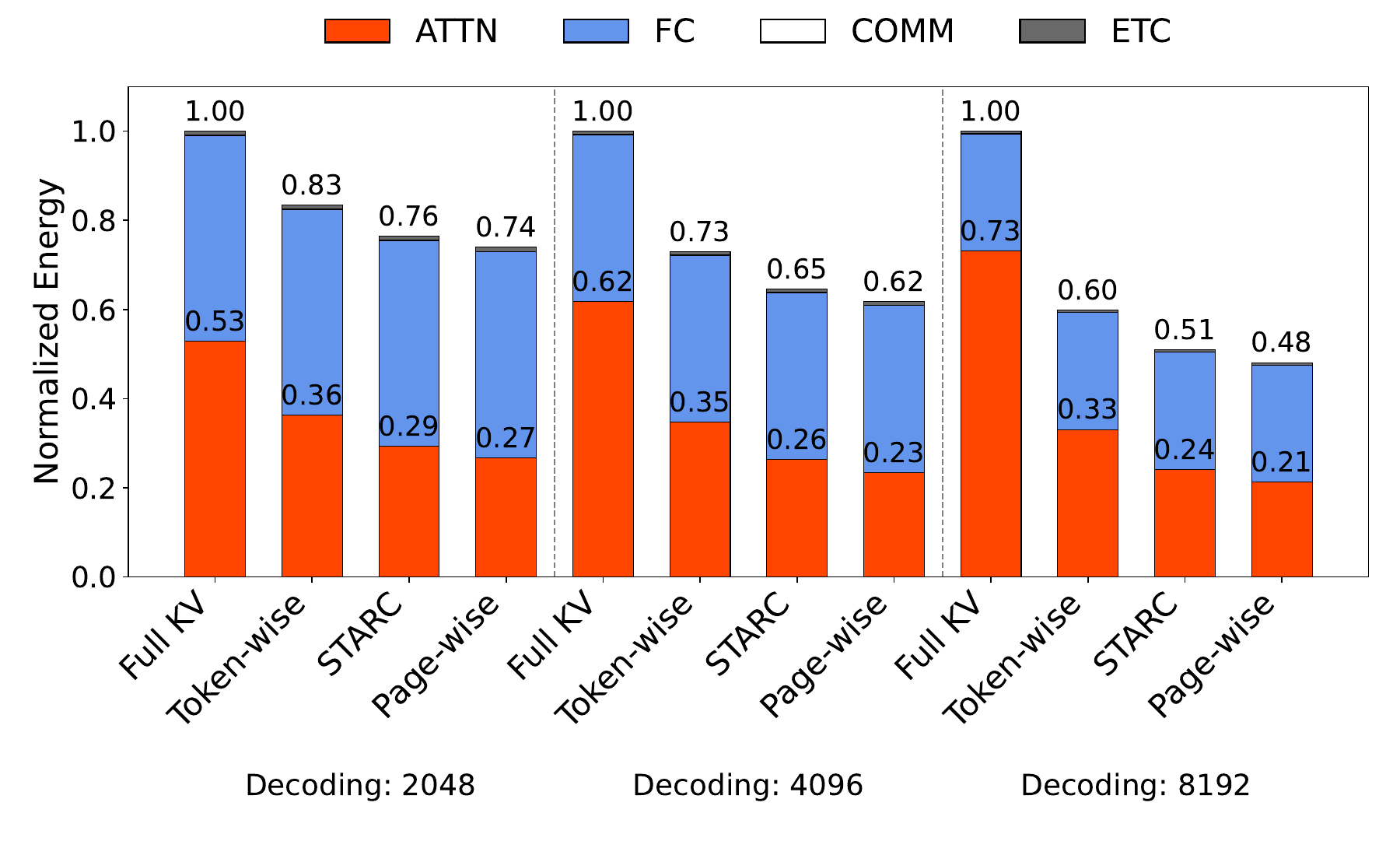} \vspace{-10pt}
  \caption{The normalized energy per output token of LLAMA 7B for various decoding length.}
  \label{fig:Energy}
\end{figure}

Fig.~\ref{fig:Energy} reports the normalized energy consumption per output token. The trends are consistent with execution time: as decoding length increases, attention energy dominates total system consumption. Token-wise sparsity also yields notable improvements over the full KV baseline. STARC further reduces attention-layer energy by 19\%--27\% compared to token-wise sparsity, and achieves up to 45\%--67\% energy reduction compared to full KV retrieval. Overall, STARC reduces total per-token energy consumption by 8\%--15\%.

While page-wise sparsity exhibits ideal alignment with the memory layout of PIM systems and achieves the best latency and energy efficiency, this comes at the expense of significantly reduced model accuracy (as shown in Section~\ref{sec:accuracy}). STARC offers a better trade-off by reducing redundant computation while preserving high retrieval quality and model accuracy, demonstrating its practical advantage for deploying sparse attention on PIM architectures.

\section{Related Work}

\label{sec:relate}

\subsection{PIM-enabled LLM Accelerators}

PIM has emerged as an effective architectural paradigm to overcome the bandwidth bottlenecks in LLMs, particularly during the autoregressive decoding stage. By placing computational units near memory arrays, PIM significantly enhances bandwidth utilization and parallelism for memory-intensive workloads. This has motivated numerous recent efforts to integrate PIM into LLM acceleration pipelines~\cite{neupims,transpim,attacc,papi,lol,cho2021accelerating}, especially for memory-bound components such as attention and feed-forward layers.

To mitigate compute-bound limitations, hybrid xPU–PIM designs have been proposed. For example, AttAcc~\cite{attacc} maps attention layers to HBM-based PIM while retaining feed-forward computation on GPUs, using pipelined execution to improve overall utilization. NeuPIMs~\cite{neupims} combines NPUs (for GEMM) and PIMs (for GEMV), using dual-row buffers and sub-batch interleaving to reduce contention. PAPI~\cite{papi} extends this model by dynamically scheduling workloads between GPUs and PIM units based on runtime profiling. However, none of these designs account for the irregular memory access patterns introduced by sparse attention.

Alongside these hybrid approaches, TransPIM~\cite{transpim} improves Transformer inference via token‑based dataflows and lightweight hardware extensions to HBM, yet it remains primarily optimized for dense computation. LoL-PIM~\cite{lol} supports long-context LLMs with a distributed PIM design and dynamic memory management, but its retrieval mechanism is agnostic to token relevance and sparsity. PIM‑LLM~\cite{pimllm} accelerates 1‑bit LLMs by using analog PIM crossbars to perform binary projection matrix multiplications and digital systolic arrays to execute 8‑bit attention matrix multiplications, yet it still assumes dense, fixed access patterns. Hermes~\cite{Hermes} leverages near-data processing DIMMs to offload cold neurons in activation-heavy workloads, focusing on activation sparsity rather than attention sparsity and lacking support for fine‑grained token selection.

In summary, most existing PIM-enabled LLM accelerators are tailored for dense and regular attention patterns, and overlook the unique challenges posed by sparse attention, such as irregular token access, dynamic KV reuse, and fine-grained selection. These issues lead to workload imbalance and inefficient memory utilization on PIM hardware. In contrast, our work introduces a sparsity-aware co-design of both memory layout and access strategy, enabling efficient execution of sparse attention under PIM architectures.

\subsection{Efficient LLM Inference}

Sparsity-based methods have been widely explored to reduce the inference cost of LLMs, particularly under long-context scenarios where the KV cache becomes a memory and latency bottleneck. A common approach is KV eviction, which permanently discards less important tokens to reduce the memory footprint. Techniques such as H2O~\cite{h2o}, Scissorhands~\cite{scissorhands} rely on ranking tokens by cumulative attention scores or recency, while FastGen~\cite{FastGen} introduces head-specific strategies for token selection. Although these methods demonstrate high compression ratio, their irreversible nature results in the loss of crucial information, as previously evicted tokens may become relevant again during decoding.

To avoid this, other methods keep the full KV cache but use dynamic sparse attention to load only the relevant tokens at runtime. For example, SparQ~\cite{sparq} approximates attention scores using query-key projections to reduce memory transfers. InfiniGen~\cite{infinigen} uses partial attention simulation to predict which tokens to prefetch. While these approaches improve bandwidth efficiency, they often overlook the architectural constraints of emerging memory systems like PIM.


To make sparse attention more hardware-friendly, Quest~\cite{quest} proposes a page-level selection strategy using query-aware scoring to identify critical KV pages, aligning better with PIM memory layouts and reducing the overhead introduced by fine-grained sparsity metadata. However, since pages are divided by token position, many selected tokens may still be irrelevant, leading to inefficiency and reduced model accuracy.

To address this, clustering-based methods improve granularity and semantic relevance in KV selection. ClusterKV~\cite{clusterkv} groups tokens by key similarity and performs cluster-level selection, capturing fine-grained attention patterns. Squeezed Attention~\cite{squeezed} applies offline clustering on fixed context segments and retrieves important clusters via centroid comparison during inference. While these methods achieve high-quality token recall within fixed computational budgets, they primarily optimize algorithmic efficiency and overlook the impact of memory system behavior.

Our method STARC builds on this line of work by co-designing clustering-based sparsity with memory-aware layout for PIM systems. Through co-design, we offer a balanced solution that enhances both model accuracy and hardware efficiency for long-context inference. 

\section{Conclusion}
In this work, we propose STARC, a clustering-based data mapping strategy that enables efficient sparse attention execution on PIM architectures. By co-locating semantically similar KV pairs and remapping them to contiguous memory regions, STARC bridges the gap between dynamic token-wise sparsity and the rigid row-level access granularity of PIM. This co-design improves both throughput and energy efficiency without compromising model accuracy. Experiments show that STARC achieves up to 31\% latency reduction and 27\% energy savings compared to token-wise sparsity baselines. We hope that our work inspires further integration of PIM architectures with emerging LLM optimization techniques, ultimately enabling scalable and efficient LLM inference in real-world deployments.

\bibliographystyle{IEEEtran}
\bibliography{main}

@inproceedings{neupims,
  title={Neupims: Npu-pim heterogeneous acceleration for batched llm inferencing},
  author={Heo, Guseul and Lee, Sangyeop and Cho, Jaehong and Choi, Hyunmin and Lee, Sanghyeon and Ham, Hyungkyu and Kim, Gwangsun and Mahajan, Divya and Park, Jongse},
  booktitle={Proceedings of the 29th ACM International Conference on Architectural Support for Programming Languages and Operating Systems, Volume 3},
  pages={722--737},
  year={2024}
}

@article{papi,
  title={PAPI: Exploiting Dynamic Parallelism in Large Language Model Decoding with a Processing-In-Memory-Enabled Computing System},
  author={He, Yintao and Mao, Haiyu and Giannoula, Christina and Sadrosadati, Mohammad and G{\'o}mez-Luna, Juan and Li, Huawei and Li, Xiaowei and Wang, Ying and Mutlu, Onur},
  journal={arXiv preprint arXiv:2502.15470},
  year={2025}
}

@inproceedings{attacc,
  title={AttAcc! Unleashing the power of PIM for batched transformer-based generative model inference},
  author={Park, Jaehyun and Choi, Jaewan and Kyung, Kwanhee and Kim, Michael Jaemin and Kwon, Yongsuk and Kim, Nam Sung and Ahn, Jung Ho},
  booktitle={Proceedings of the 29th ACM International Conference on Architectural Support for Programming Languages and Operating Systems, Volume 2},
  pages={103--119},
  year={2024}
}

@inproceedings{transpim,
  title={TransPIM: A memory-based acceleration via software-hardware co-design for transformer},
  author={Zhou, Minxuan and Xu, Weihong and Kang, Jaeyoung and Rosing, Tajana},
  booktitle={2022 IEEE International Symposium on High-Performance Computer Architecture (HPCA)},
  pages={1071--1085},
  year={2022},
  organization={IEEE}
}

@inproceedings{infinigen,
  title={$\{$InfiniGen$\}$: Efficient generative inference of large language models with dynamic $\{$KV$\}$ cache management},
  author={Lee, Wonbeom and Lee, Jungi and Seo, Junghwan and Sim, Jaewoong},
  booktitle={18th USENIX Symposium on Operating Systems Design and Implementation (OSDI 24)},
  pages={155--172},
  year={2024}
}

@article{sparq,
  title={Sparq attention: Bandwidth-efficient llm inference},
  author={Ribar, Luka and Chelombiev, Ivan and Hudlass-Galley, Luke and Blake, Charlie and Luschi, Carlo and Orr, Douglas},
  journal={arXiv preprint arXiv:2312.04985},
  year={2023}
}

@article{model,
  title={Model tells you where to merge: Adaptive kv cache merging for llms on long-context tasks},
  author={Wang, Zheng and Jin, Boxiao and Yu, Zhongzhi and Zhang, Minjia},
  journal={arXiv preprint arXiv:2407.08454},
  year={2024}
}

@article{clusterkv,
  title={Clusterkv: Manipulating llm kv cache in semantic space for recallable compression},
  author={Liu, Guangda and Li, Chengwei and Zhao, Jieru and Zhang, Chenqi and Guo, Minyi},
  journal={arXiv preprint arXiv:2412.03213},
  year={2024}
}

@article{lol,
  title={LoL-PIM: Long-Context LLM Decoding with Scalable DRAM-PIM System},
  author={Kwon, Hyucksung and Koo, Kyungmo and Kim, Janghyeon and Lee, Woongkyu and Lee, Minjae and Lee, Hyungdeok and Jung, Yousub and Park, Jaehan and Song, Yosub and Yang, Byeongsu and others},
  journal={arXiv preprint arXiv:2412.20166},
  year={2024}
}

@article{quest,
  title={Quest: Query-aware sparsity for efficient long-context llm inference},
  author={Tang, Jiaming and Zhao, Yilong and Zhu, Kan and Xiao, Guangxuan and Kasikci, Baris and Han, Song},
  journal={arXiv preprint arXiv:2406.10774},
  year={2024}
}

@article{squeezed,
  title={Squeezed attention: Accelerating long context length llm inference},
  author={Hooper, Coleman and Kim, Sehoon and Mohammadzadeh, Hiva and Maheswaran, Monishwaran and Paik, June and Mahoney, Michael W and Keutzer, Kurt and Gholami, Amir},
  journal={arXiv preprint arXiv:2411.09688},
  year={2024}
}

@article{pimllm,
  title={PIM-LLM: A High-Throughput Hybrid PIM Architecture for 1-bit LLMs},
  author={Malekar, Jinendra and Chandarana, Peyton and Amin, Md Hasibul and Elbtity, Mohammed E and Zand, Ramtin},
  journal={arXiv preprint arXiv:2504.01994},
  year={2025}
}

@inproceedings{Hermes,
  title={Make LLM Inference Affordable to Everyone: Augmenting GPU Memory with NDP-DIMM},
  author={Liu, Lian and Zhao, Shixin and Li, Bing and Ren, Haimeng and Xu, Zhaohui and Wang, Mengdi and Li, Xiaowei and Han, Yinhe and Wang, Ying},
  booktitle={2025 IEEE International Symposium on High Performance Computer Architecture (HPCA)},
  pages={1751--1765},
  year={2025},
  organization={IEEE}
}

@article{h2o,
  title={H2o: Heavy-hitter oracle for efficient generative inference of large language models},
  author={Zhang, Zhenyu and Sheng, Ying and Zhou, Tianyi and Chen, Tianlong and Zheng, Lianmin and Cai, Ruisi and Song, Zhao and Tian, Yuandong and R{\'e}, Christopher and Barrett, Clark and others},
  journal={Advances in Neural Information Processing Systems},
  volume={36},
  pages={34661--34710},
  year={2023}
}

@article{scissorhands,
  title={Scissorhands: Exploiting the persistence of importance hypothesis for llm kv cache compression at test time},
  author={Liu, Zichang and Desai, Aditya and Liao, Fangshuo and Wang, Weitao and Xie, Victor and Xu, Zhaozhuo and Kyrillidis, Anastasios and Shrivastava, Anshumali},
  journal={Advances in Neural Information Processing Systems},
  volume={36},
  pages={52342--52364},
  year={2023}
}

@article{FastGen,
  title={Model tells you what to discard: Adaptive kv cache compression for llms},
  author={Ge, Suyu and Zhang, Yunan and Liu, Liyuan and Zhang, Minjia and Han, Jiawei and Gao, Jianfeng},
  journal={arXiv preprint arXiv:2310.01801},
  year={2023}
}

@inproceedings{newton,
  title={Newton: A DRAM-maker’s accelerator-in-memory (AiM) architecture for machine learning},
  author={He, Mingxuan and Song, Choungki and Kim, Ilkon and Jeong, Chunseok and Kim, Seho and Park, Il and Thottethodi, Mithuna and Vijaykumar, TN},
  booktitle={2020 53rd Annual IEEE/ACM International Symposium on Microarchitecture (MICRO)},
  pages={372--385},
  year={2020},
  organization={IEEE}
}

@inproceedings{bai2024longbench,
    title = "{L}ong{B}ench: A Bilingual, Multitask Benchmark for Long Context Understanding",
    author = "Bai, Yushi and Lv, Xin  and Zhang, Jiajie  and Lyu, Hongchang  and
      Tang, Jiankai  and Huang, Zhidian  and Du, Zhengxiao  and Liu, Xiao  and Zeng, Aohan  and Hou, Lei  and Dong, Yuxiao  and Tang, Jie  and Li, Juanzi",
    booktitle = "Proceedings of the 62nd Annual Meeting of the Association for Computational Linguistics (Volume 1: Long Papers)",
    month = aug,
    year = "2024",
    address = "Bangkok, Thailand",
    publisher = "Association for Computational Linguistics",
    url = "https://aclanthology.org/2024.acl-long.172",
    doi = "10.18653/v1/2024.acl-long.172",
    pages = "3119--3137",
}

@article{rae2019compressive,
  title={Compressive transformers for long-range sequence modelling},
  author={Rae, Jack W and Potapenko, Anna and Jayakumar, Siddhant M and Lillicrap, Timothy P},
  journal={arXiv preprint arXiv:1911.05507},
  year={2019}
}

@article{achiam2023gpt,
  title={Gpt-4 technical report},
  author={Achiam, Josh and Adler, Steven and Agarwal, Sandhini and Ahmad, Lama and Akkaya, Ilge and Aleman, Florencia Leoni and Almeida, Diogo and Altenschmidt, Janko and Altman, Sam and Anadkat, Shyamal and others},
  journal={arXiv preprint arXiv:2303.08774},
  year={2023}
}

@article{zhuang2024toolqa,
  title={Toolqa: A dataset for llm question answering with external tools},
  author={Zhuang, Yuchen and Yu, Yue and Wang, Kuan and Sun, Haotian and Zhang, Chao},
  journal={Advances in Neural Information Processing Systems},
  volume={36},
  year={2024}
}

@inproceedings{svyatkovskiy2019pythia,
  title={Pythia: Ai-assisted code completion system},
  author={Svyatkovskiy, Alexey and Zhao, Ying and Fu, Shengyu and Sundaresan, Neel},
  booktitle={Proceedings of the 25th ACM SIGKDD international conference on knowledge discovery \& data mining},
  pages={2727--2735},
  year={2019}
}

@article{roziere2023code,
  title={Code llama: Open foundation models for code},
  author={Roziere, Baptiste and Gehring, Jonas and Gloeckle, Fabian and Sootla, Sten and Gat, Itai and Tan, Xiaoqing Ellen and Adi, Yossi and Liu, Jingyu and Remez, Tal and Rapin, J{\'e}r{\'e}my and others},
  journal={arXiv preprint arXiv:2308.12950},
  year={2023}
}

@article{shinn2023reflexion,
  title={Reflexion: Language agents with verbal reinforcement learning},
  author={Shinn, Noah and Cassano, Federico and Gopinath, Ashwin and Narasimhan, Karthik and Yao, Shunyu},
  journal={Advances in Neural Information Processing Systems},
  volume={36},
  pages={8634--8652},
  year={2023}
}

@article{yao2023tree,
  title={Tree of thoughts: Deliberate problem solving with large language models},
  author={Yao, Shunyu and Yu, Dian and Zhao, Jeffrey and Shafran, Izhak and Griffiths, Tom and Cao, Yuan and Narasimhan, Karthik},
  journal={Advances in neural information processing systems},
  volume={36},
  pages={11809--11822},
  year={2023}
}

@article{pope2023efficiently,
  title={Efficiently scaling transformer inference},
  author={Pope, Reiner and Douglas, Sholto and Chowdhery, Aakanksha and Devlin, Jacob and Bradbury, James and Heek, Jonathan and Xiao, Kefan and Agrawal, Shivani and Dean, Jeff},
  journal={Proceedings of Machine Learning and Systems},
  volume={5},
  pages={606--624},
  year={2023}
}

@inproceedings{kwon2023efficient,
  title={Efficient memory management for large language model serving with pagedattention},
  author={Kwon, Woosuk and Li, Zhuohan and Zhuang, Siyuan and Sheng, Ying and Zheng, Lianmin and Yu, Cody Hao and Gonzalez, Joseph and Zhang, Hao and Stoica, Ion},
  booktitle={Proceedings of the 29th Symposium on Operating Systems Principles},
  pages={611--626},
  year={2023}
}

@article{li2022pre,
  title={Pre-trained language models for interactive decision-making},
  author={Li, Shuang and Puig, Xavier and Paxton, Chris and Du, Yilun and Wang, Clinton and Fan, Linxi and Chen, Tao and Huang, De-An and Aky{\"u}rek, Ekin and Anandkumar, Anima and others},
  journal={Advances in Neural Information Processing Systems},
  volume={35},
  pages={31199--31212},
  year={2022}
}

@inproceedings{gao2019computedram,
  title={Computedram: In-memory compute using off-the-shelf drams},
  author={Gao, Fei and Tziantzioulis, Georgios and Wentzlaff, David},
  booktitle={Proceedings of the 52nd annual IEEE/ACM international symposium on microarchitecture},
  pages={100--113},
  year={2019}
}

@inproceedings{hyun2024pathfinding,
  title={Pathfinding future pim architectures by demystifying a commercial pim technology},
  author={Hyun, Bongjoon and Kim, Taehun and Lee, Dongjae and Rhu, Minsoo},
  booktitle={2024 IEEE International Symposium on High-Performance Computer Architecture (HPCA)},
  pages={263--279},
  year={2024},
  organization={IEEE}
}

@article{oliveira2022accelerating,
  title={Accelerating neural network inference with processing-in-DRAM: from the edge to the cloud},
  author={Oliveira, Geraldo F and G{\'o}mez-Luna, Juan and Ghose, Saugata and Boroumand, Amirali and Mutlu, Onur},
  journal={IEEE Micro},
  volume={42},
  number={6},
  pages={25--38},
  year={2022},
  publisher={IEEE}
}

@inproceedings{cho2021accelerating,
  title={Accelerating bandwidth-bound deep learning inference with main-memory accelerators},
  author={Cho, Benjamin Y and Jung, Jeageun and Erez, Mattan},
  booktitle={Proceedings of the International Conference for High Performance Computing, Networking, Storage and Analysis},
  pages={1--14},
  year={2021}
}

@inproceedings{imani2019floatpim,
  title={Floatpim: In-memory acceleration of deep neural network training with high precision},
  author={Imani, Mohsen and Gupta, Saransh and Kim, Yeseong and Rosing, Tajana},
  booktitle={Proceedings of the 46th International Symposium on Computer Architecture},
  pages={802--815},
  year={2019}
}

@misc{longchat2023,
    title = {How Long Can Open-Source LLMs Truly Promise on Context Length?},
    url = {https://lmsys.org/blog/2023-06-29-longchat},
    author = {Dacheng Li and Rulin Shao and Anze Xie and Ying Sheng and Lianmin Zheng and Joseph E. Gonzalez and Ion Stoica and Xuezhe Ma and Hao Zhang},
    month = {June},
    year = {2023}
}

@article{yang2018hotpotqa,
  title={HotpotQA: A dataset for diverse, explainable multi-hop question answering},
  author={Yang, Zhilin and Qi, Peng and Zhang, Saizheng and Bengio, Yoshua and Cohen, William W and Salakhutdinov, Ruslan and Manning, Christopher D},
  journal={arXiv preprint arXiv:1809.09600},
  year={2018}
}

@article{dasigi2021dataset,
  title={A dataset of information-seeking questions and answers anchored in research papers},
  author={Dasigi, Pradeep and Lo, Kyle and Beltagy, Iz and Cohan, Arman and Smith, Noah A and Gardner, Matt},
  journal={arXiv preprint arXiv:2105.03011},
  year={2021}
}

@article{kovcisky2018narrativeqa,
  title={The narrativeqa reading comprehension challenge},
  author={Ko{\v{c}}isk{\`y}, Tom{\'a}{\v{s}} and Schwarz, Jonathan and Blunsom, Phil and Dyer, Chris and Hermann, Karl Moritz and Melis, G{\'a}bor and Grefenstette, Edward},
  journal={Transactions of the Association for Computational Linguistics},
  volume={6},
  pages={317--328},
  year={2018},
  publisher={MIT Press One Rogers Street, Cambridge, MA 02142-1209, USA journals-info~…}
}

@article{joshi2017triviaqa,
  title={Triviaqa: A large scale distantly supervised challenge dataset for reading comprehension},
  author={Joshi, Mandar and Choi, Eunsol and Weld, Daniel S and Zettlemoyer, Luke},
  journal={arXiv preprint arXiv:1705.03551},
  year={2017}
}

@article{gliwa2019samsum,
  title={SAMSum corpus: A human-annotated dialogue dataset for abstractive summarization},
  author={Gliwa, Bogdan and Mochol, Iwona and Biesek, Maciej and Wawer, Aleksander},
  journal={arXiv preprint arXiv:1911.12237},
  year={2019}
}

@inproceedings{guo2023longcoder,
  title={Longcoder: A long-range pre-trained language model for code completion},
  author={Guo, Daya and Xu, Canwen and Duan, Nan and Yin, Jian and McAuley, Julian},
  booktitle={International Conference on Machine Learning},
  pages={12098--12107},
  year={2023},
  organization={PMLR}
}

@article{raffel2020exploring,
  title={Exploring the limits of transfer learning with a unified text-to-text transformer},
  author={Raffel, Colin and Shazeer, Noam and Roberts, Adam and Lee, Katherine and Narang, Sharan and Matena, Michael and Zhou, Yanqi and Li, Wei and Liu, Peter J},
  journal={Journal of machine learning research},
  volume={21},
  number={140},
  pages={1--67},
  year={2020}
}

\end{document}